\documentclass[conference]{IEEEtran}
\IEEEoverridecommandlockouts

\usepackage{cite}
\usepackage{amsmath,amssymb,amsfonts}
\usepackage{algorithmic}
\usepackage{graphicx}
\usepackage{textcomp}
\usepackage{xcolor}
\usepackage{multirow}
\def\BibTeX{{\rm B\kern-.05em{\sc i\kern-.025em b}\kern-.08em
    T\kern-.1667em\lower.7ex\hbox{E}\kern-.125emX}}
\begin{document}

\title{Unified Face Attack Detection via Fine-Grained Semantic Guidance}

\author{\IEEEauthorblockN{Ning Jiang\textsuperscript{1},
Shijie Yu\textsuperscript{2},
Dingheng Zeng\textsuperscript{2}, Haiyang Yi\textsuperscript{2}, Yanhong Liu\textsuperscript{2}, Haifeng Shen\textsuperscript{2} and
Ying Li\textsuperscript{1}\IEEEauthorrefmark{1}}
\IEEEauthorblockA{\textsuperscript{1}School of Software \& Microelectronics, Peking University, Beijing, China}
\IEEEauthorblockA{\textsuperscript{2}Mashang Consumer Finance Co., Ltd., Chongqing, China}
\thanks{ 
\IEEEauthorrefmark{1} indicates corresponding author.}
}

\maketitle

\begin{abstract}
The growing applications of facial recognition systems are accompanied by increasingly diverse security threats. Existing datasets lack detailed textual descriptions of forgery cues, leading most prior methods to treat face attack detection primarily as a visual recognition task. In this paper, building upon the large-scale MS-UFAD dataset which contains over 8 million attack images, we enrich each image with a fine-grained textual description of forgery cues. Furthermore, we propose a \textit{Dual Alignment Forgery Network} (DAF-Net) to better leverage these textual information. Extensive experiments demonstrate that our approach extracts more generalizable and semantically meaningful forgery representations from attack images, outperforming both vision-only methods and approaches based on coarse-grained descriptions.
\end{abstract}


\section{Introduction}
\label{sec:intro}
As the significant development of generative AI techniques, face manipulation and generation is widely used in entertainment. However, these techniques can also be misused to hack facial recognition systems for malicious purposes, such as the creation of deepfakes. 
Additionally, facial recognition systems face threats from adversarial attacks, and spoofing techniques like print-attacks or replay-attacks. 
Consequently, developing reliable and robust defense mechanisms to counteract diverse real-world facial attacks has become critically urgent.

Previous research has developed various methods to defend against such attacks. Most existing approaches \cite{fang2024unified,deb,cao2022end,li2020face} rely solely on visual information, primarily because existing datasets lack textual descriptions. Although recent studies \cite{fang2024vl,zhang2025mfclip,wang2024tf} have attempted to incorporate textual descriptions, they typically constructed class-level texts (\textit{e.g}, \textit{“This is a fake face.”}), which lack fine-grained detail. In 2025, Jiang et al.\cite{ms-ufad} introduced MS-UFAD, a large-scale face attack dataset which includes textual descriptions of artifacts in each video, but a limitation persists: for instance, talking-face videos generated from the same source video but manipulated with different methods are assigned the same description, even though the visual artifacts differ significantly. This can result in inaccurate or misleading guidance.

\begin{figure}
    \centering
    \includegraphics[width=0.85\linewidth]{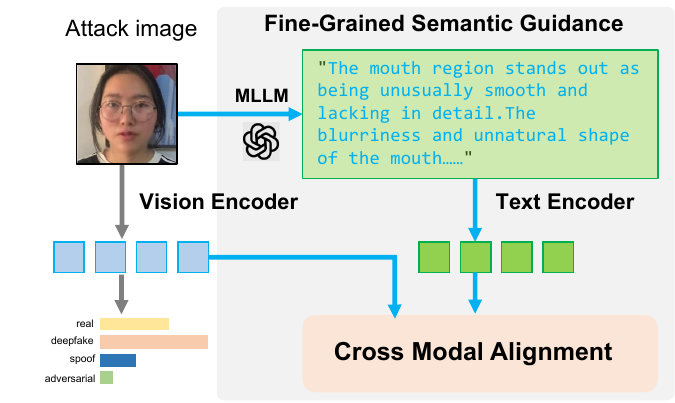}
    \caption{An overview of face attack detection via fine-grained semantic guidance. The guidance part will be removed during inference.}
    \label{fig:teaser}
\end{figure}

To address this issue, as shown in Fig. \ref{fig:teaser}, we propose to employ fine-grained descriptions to enhance the generalization of face attack detection models. Firstly, we enrich the MS-UFAD dataset by supplementing each image with fine-grained textual descriptions of forgery cues. Specifically, we employ multimodal large language models (MLLMs) to generate these descriptions via carefully designed prompts tailored to different attack types. In total, the enriched dataset comprises over 8 million textual annotations. To fully leverage these fine‑grained descriptions, we propose a \textit{Dual Alignment Forgery Network} (DAF‑Net), which consists of two symmetric sub‑networks for encoding face images and text, respectively. During inference, only the visual branch is utilized. Additionally, we design a lightweight module named \textit{Semantic Forgery Aggregation Module} (SFAM) to aggregate discrete visual patches and textual tokens into semantic meaningful forgery regions and phrases. This module, which consists of multiple stacked transformer blocks equipped with cross‑attention mechanisms, employs $N$ learnable query vectors to aggregate forgery-related tokens from the corresponding encoders. To enhance multi‑granular interaction between the two modalities, we apply both a global and a fine‑grained alignment loss. Extensive experiments demonstrate that with the aid of fine-grained textual descriptions, DAF-Net can learn more generalizable and semantically meaningful forgery representations from attack images, outperforming both vision-only methods and those based on coarse-grained textual descriptions.

Our contributions are summarized as follows:
\begin{itemize}
    \item We enrich the MS-UFAD dataset by incorporating more than 8 million fine‑grained textual annotations that describe forgery cues within attack images, establishing a large‑scale multi-modal resource to advance research in face attack detection.
    \item We propose the Dual Alignment Forgery Network along with Semantic Forgery Aggregation modules, which effectively leverages fine‑grained textual descriptions to learn generalizable and semantical forgery features.
    \item Extensive experimental evaluations confirm the efficacy of fine‑grained textual descriptions for face attack detection. Our method consistently outperforms not only vision‑only methods but also approaches that rely on coarse‑grained descriptions.
\end{itemize}

\begin{figure}
    \centering
    \includegraphics[width=0.9\linewidth]{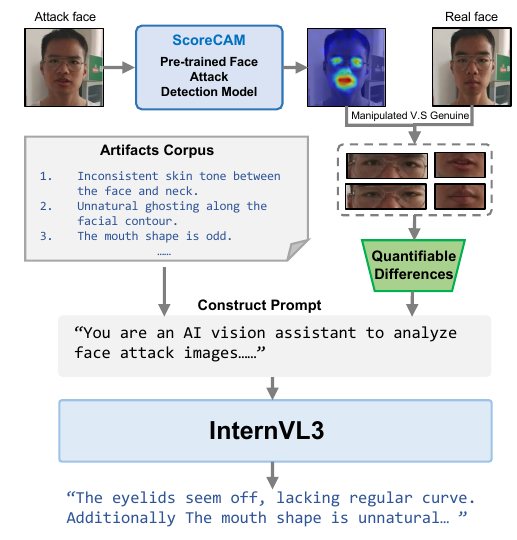}
    \caption{The pipeline for generating descriptions. Initially, ScoreCAM\cite{wang2020score} is employed to localize manipulated regions using a pre-trained model. These regions are then compared with corresponding genuine face images to quantify differences in attributes such as color and blur. The prompt built upon the differences and artifact corpus guides InternVL3\cite{zhu2025internvl3exploringadvancedtraining} to produce a detailed description of the forgery cues present in the attack image.}
    \label{fig:datapipe}
\end{figure}

\section{Related Work}

\textbf{Face Attack Dataset.} Most existing attack datasets, including  FaceForensics++ \cite{rossler2019faceforensics++}, Celeb-DF \cite{Celeb_DF_cvpr20}, DFDC \cite{dolhansky2020deepfake}, OULU-NPU \cite{oulu-npu}, CASIA-SURF \cite{zhang2020casia}, CelebA-Spoof \cite{zhang2020celeba}, GrandFake \cite{deb} and UniAttackData \cite{fang2024unified}, primarily provide attack types, lacking textual descriptions of forgery cues in attack images/videos. A few recent attempts \cite{Sun_2025_CVPR, ms-ufad,  zhang2024common} have explored generating descriptions by human annotation or MLLMs, such as DD-VQA \cite{zhang2024common} and MS-UFAD \cite{ms-ufad} yet such captions are limited in scale, noisy, and lack the fidelity required to describe fine-grained forensic cues.

\textbf{Face Attack Detection Methods.} Early face attack detection \cite{boulkenafet2015face, komulainen2013context, patel2016secure} research primarily focused on handcrafted features  extracting texture or motion patterns to distinguish bonafide from presentation attacks (e.g., printed photos or 3D masks). With the rise of deep learning, CNNs became dominant for capturing attack traces \cite{atoum2017face, rossler2019faceforensics++, shiohara2022detecting, yan2023ucf}. However, most prior work still treats forgery detection as a purely visual task, leaving models vulnerable to dataset bias and the rapid evolution of attack methods. Multi-modal methods leveraged language-guided pretraining (\textit{e.g.}, CLIP \cite{radford2021learning}) to enhance zero-shot transfer. Recent efforts \cite{zhang2025mfclip, fang2024vl, ms-ufad, guo2025rethinking} address high-fidelity threats (\textit{e.g.}, diffusion-generated faces) via noise-texture alignment or interpretable frameworks.  Nevertheless, most existing approaches rely on manually crafted prompts or generic high-level sentences (\textit{e.g.}, \textit{``this is a fake face"}), which fail to capture rich fine-grained features.

\begin{figure}
    \centering
    \includegraphics[width=0.85\linewidth]{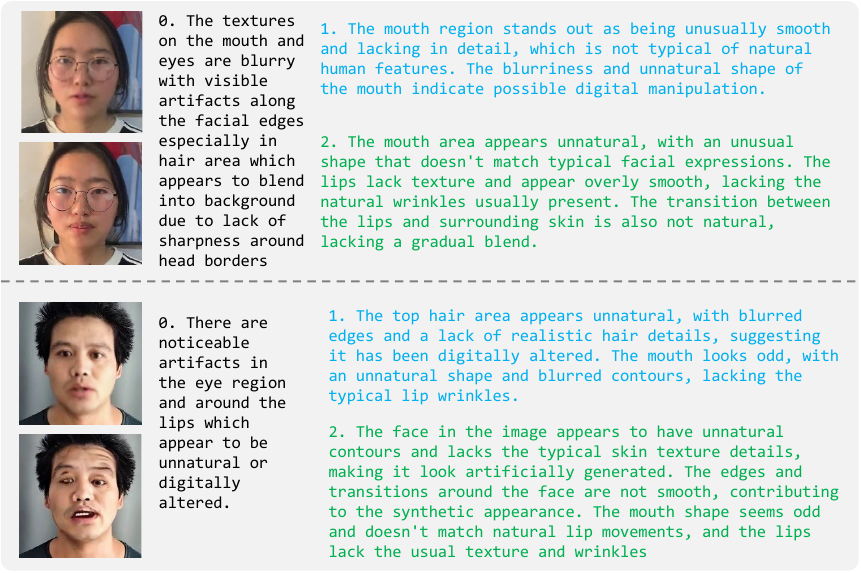}
    \caption{The comparison between the descriptions from MS-UFAD dataset (black) and our generated descriptions (blue and green).}
    \label{fig:cmp}
\end{figure}

\section{Dataset}

\begin{figure*}
    \centering
    \includegraphics[width=1.0\linewidth]{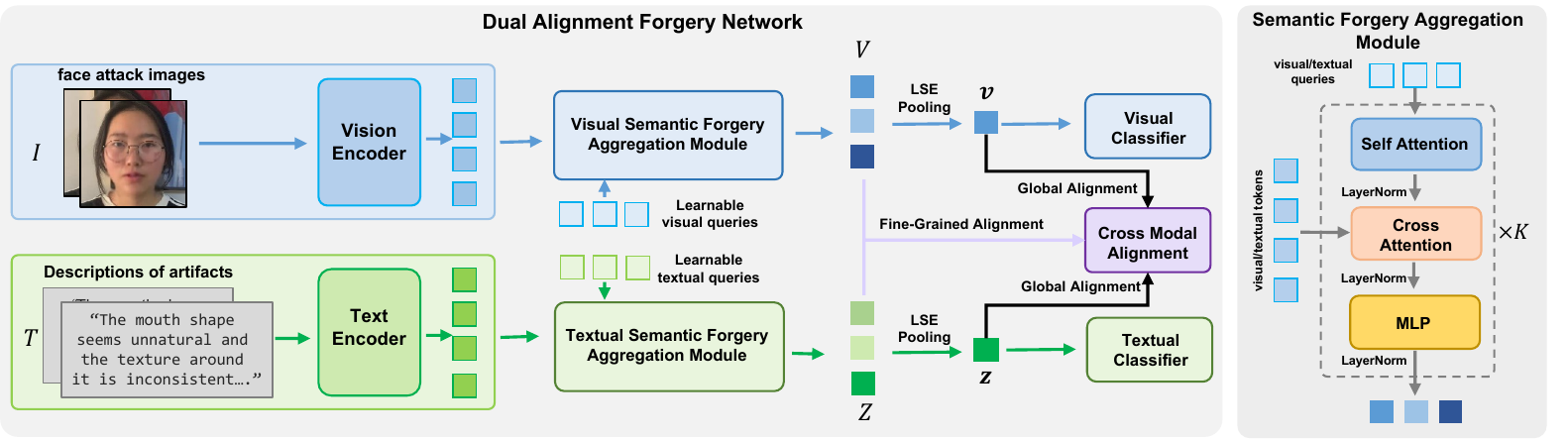}
    \caption{An overview of the proposed Dual Alignment Forgery Network (DAF-Net). The architecture takes face attack images and the corresponding textual artifact descriptions as input. A vision encoder and a text encoder process the respective inputs. The Visual and Textual Semantic Forgery Aggregation modules then extract and refine forgery-specific features. Then the aligned features are finally pooled and processed by the visual and textual classifiers to produce the final prediction. During inference, the textual branch will be removed.}
    \label{fig:network}
\end{figure*}

\subsection{Textual Description Generation}
Multimodal large language models (MLLMs) have demonstrated notable advancements not only in comprehending visual content but also in generating contextually relevant text. Building on this, we employ MLLMs to generate detailed descriptions of the forgery traces in face attack images. However, the hallucination issue inherent in MLLMs remains a significant concern. When directly applied to generate descriptions of forgery cues, two primary limitations arise: 1) Imprecise localization of manipulated regions, and 2) Semantic misalignment between the generated descriptions and the actual forgery traces present in the image. These limitations can compromise the reliability of the generated textual descriptions. 

To address this, we developed a novel pipeline for generating descriptions as shown in Fig. \ref{fig:datapipe}. Specifically, we begin with a pre-trained face attack detection model. We then apply the ScoreCAM\cite{wang2020score} algorithm to this model to produce visual explanations in the form of heatmaps. Within these heatmaps, regions of high activation (significantly brighter areas) are interpreted as indicators of the manipulated regions. Then, we crop the regions from both the attack image and the corresponding real face. Subsequently, inspired by \cite{Sun_2025_CVPR}, we calculate the quantifiable difference of color/blur between the manipulated regions and genuine regions to obtain more accurate description (\textit{e.g.}, \textit{slightly/highly blurred}). However, some discrepancies cannot be fully quantified, such as the implausible facial contour warping or unnatural physiological textures. Therefore, we develop distinct artifact corpus tailored to specific forgery types. For instance, in the case of FaceSwap, the corpus includes descriptors such as "\textit{Inconsistent skin tone between the face and neck}" and "\textit{Unnatural ghosting along the facial contour}", \textit{etc}. These curated descriptions are incorporated into the prompt, enabling MLLMs (InternVL3 \cite{zhu2025internvl3exploringadvancedtraining} is used here) to select the most related descriptions based on the input attack image. Subsequently, the MLLMs synthesize the quantifiable discrepancies and the chosen artifact descriptions into coherent and fluent natural language output.

As the two examples shown in Fig. \ref{fig:cmp}, the descriptions from MS-UFAD dataset (marked with black) have obvious weakness that the two attack images of the same individual, produced by different generative algorithms, are assigned the same textual description  despite exhibiting noticeable visual differences. In contrast, our approach provides a distinct and tailored description for each image (marked with blue and green). Moreover, the descriptions we generate are also more accurate and detailed.

\subsection{Dataset Protocol}
Statistically, the extensive MS-UFAD dataset comprises approximately 8 million images alongside corresponding textual descriptions of forgery artifacts. This data is drawn from 5,000 unique individuals and is partitioned according to both individual identity and the specific generative method used. To increase the task's difficulty, the training set is deliberately limited, containing only 830,000 images and descriptions from 2,000 individuals, generated by just 10 methods. The test set, in contrast, encompasses the remaining data, which spans 30 generative methods. Due to the substantial size of the test set (over 3.6 million items), a representative subset of 201,453 image-text pairs is sampled, balancing both person identity and generative method, for final evaluation.

\section{Methodology}

\subsection{Network Architecture}
To fully exploit the generalizable forgery representations from attack images with the aid of fine-grained textual descriptions, this paper proposes a \textit{Dual Alignment Forgery Network} (DAF-Net) which consists of visual and textual branches. Specifically, the backbone employs the visual and textual encoders from the BLIP model \cite{li2022blip}, while a \textit{Semantic Forgery Aggregation Module} (SFAM, see Section \ref{SFAM} for more details) is inserted into each branch. It should be noted that only the visual branch is utilized during inference. The overall framework of DAF-Net is shown in Fig. \ref{fig:network}.

Given a mini-batch of attack face images $\mathcal{I}$ and the corresponding textual descriptions of forgery cues $\mathcal{T}$, the visual and textual encoders of DAF-Net (denoted as $\mathit{E_v}(\cdot)$ and $\mathit{E_t}(\cdot)$, respectively) first encode them into visual patch embeddings $\mathit{E_v}(\mathcal{I}) \in \mathbb{R}^{B\times L_v \times D}$ and textual token embeddings $\mathit{E_t}(\mathcal{T}) \in \mathbb{R}^{B\times L_t \times D}$. Subsequently, the SFAM equipped cross-attention mechanism employs $N$ learnable query vectors to aggregate discrete visual patches and textual tokens into semantic meaningful forgery regions and phrases, producing outputs $\mathit{V} \in \mathbb{R}^{B\times N \times D}$ and $\mathit{Z} \in \mathbb{R}^{B\times N \times D}$ respectively. To further aggregate the feature representations to one vector, Log-Sum-Exp (LSE) pooling is employed. This pooling operation emphasizes more salient forgery features while preserving fine‑grained information. The visual representation $\mathbf{v} \in \mathbb{R}^{B\times D}$ is obtained by
$
    \mathbf{v}^j_k = \frac{1}{r}\log\big(\frac{1}{N}\sum_{i=1}^{N}\exp(r\cdot V_{k}^{ij})\big)
$
where $\mathbf{v}^j_k$ denote the $j$-th element of the visual representation of $k$-th sample in a mini-batch, and $r$ is a scaling parameter balancing the pooling behavior. The textual representation $\mathbf{z}$ is derived in a similar manner. 

Finally, the visual and textual representations are used for classification at the output stage of DAF‑Net and a classification loss function is employed to supervise both branches:
\begin{equation}
    \mathcal{L}_{cls} = \mathcal{L}_{ce}(\mathcal{C}_v(\mathbf{v}), \mathbf{y}) + \mathcal{L}_{ce}(\mathcal{C}_z(\mathbf{z}), \mathbf{y})
\end{equation}
where $\mathcal{L}_{ce}$ denotes cross entropy loss function, $\mathcal{C}_{*}(\cdot)$ denotes the logits output by a classifier, and $\mathbf{y}$ denotes the ground truth labels of attack images.

\subsection{Semantic Forgery Aggregation Module}
\label{SFAM}
In an attack image, a manipulated region typically comprises multiple spatially adjacent patches, and its corresponding textual description typically forms a semantically complete phrase or sentence. However, the visual patches $\mathit{E_v}(\mathcal{I})$ produced by the visual encoder are spatially discontinuous, and a single textual token in $\mathit{E_t}(\mathcal{T})$ may not even represent a complete word. This structural misalignment poses a significant challenge to establishing fine-grained visual-textual correspondence.

To address this issue, we design a light-weight module named \textit{Semantic Forgery Aggregation Module} (SFAM). Specifically, the visual SFAM is designed to filter out forgery-unrelated visual patches and implicitly aggregate forgery-related visual patches into $N$ forgery visual regions, denoted as $\mathit{V} \in \mathbb{R}^{B\times N \times D}$. Similarly,  the textual SFAM aggregates textual tokens into $N$ forgery textual phrases $\mathit{Z} \in \mathbb{R}^{B\times N \times D}$. As shown in the right of Fig. \ref{fig:network}, each SFAM, consisting of $K$ stacked transformer blocks with cross-attention mechanism, employs $N$ learnable vectors to interact with visual patches or textual tokens. Take visual stream as an example. The $N$ learnable vectors serve as the query, while the visual patches $\mathit{E_v}(\mathcal{I})$ are regarded as the key and value. 

\subsection{Cross Modal Alignment}
\subsubsection{Global Cross Modal Alignment}
We employ contrastive loss which operates at the level of global visual and textual representation, \textit{i.e.}, $\mathbf{v}$ and $\mathbf{z}$, respectively. Specifically, 
we optimize
\begin{equation}
\begin{split}
        \mathcal{L}_{global} = -\frac{1}{2B}\sum_{i=1}^B\bigg(\log\frac{\exp({\phi(\mathbf{v}_i, \mathbf{z}_i)/\tau})}{\sum_{j=1}^{B}\exp({\phi(\mathbf{v}_i, \mathbf{z}_j) / \tau})} \\+ \log\frac{\exp({\phi(\mathbf{v}_i, \mathbf{z}_i)/\tau})}{\sum_{j=1}^{B}\exp({\phi(\mathbf{v}_j, \mathbf{z}_i) / \tau})}\bigg)
\end{split}
\end{equation}
where $\phi(\mathbf{v}_i, \mathbf{z}_j)$ denotes the cosine similarity between the $i$-th visual representation and $j$-th textual representation in a mini-batch, and  $\tau$ is the temperature parameter.

\subsubsection{Fine-Grained Cross Modal Alignment}
Global contrastive loss primarily optimizes sample-level visual-textual feature similarity, which can result in the loss of fine-grained details, particularly when the input text description is lengthy. To further enhance fine-grained visual-textual alignment, we employ Cross Modal Late Interaction\cite{yao2022filip} on $V$ and $Z$ output by the both SFAMs instead of using $\mathit{E_v}(\mathcal{I})$ and $\mathit{E_t}(\mathcal{T})$ directly.

\begin{figure}
    \centering
    \includegraphics[width=0.9\linewidth]{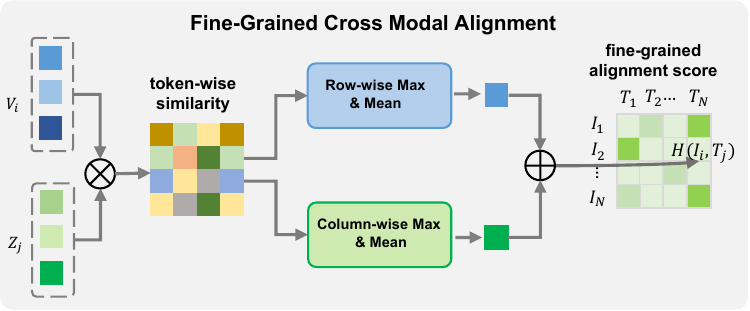}
    \caption{The pipeline of fine-grained cross modal alignment.  }
    \label{fig:alignment}
\end{figure}

As shown in Fig. \ref{fig:alignment}, given $i$-th face attack image and $j$-th textual description in a mini-batch, we first compute the token-wise similarity matrix $\mathbf{S} = \hat{V}_i\hat{Z}_j^T \in \mathbb{R}^{N\times N}$ where $\hat{\cdot}$ denotes the normalization operation. Then, we choose the best-matching textual phrase (or visual region) for each region (or each phrase), and calculate the average of these $N$ max scores to represent the final fine-grained alignment score:
\begin{equation}
    \mathcal{H}(\mathcal{I}_i, \mathcal{T}_j) = \frac{1}{N}\sum_{n=1}^N\big(\max(\mathbf{S}[n, :]) + \max(\mathbf{S}[:, n])\big)
\end{equation}
where $\max(\mathbf{S}[n, :])$ denotes the max value of $n$-th row of the token-wise similarity matrix, \textit{i.e.}, the max similarity score of $n$-th visual region to a certain textual phrase. Similarly for $\max(\mathbf{S}[:, n])$, it denotes the max similarity score of $n$-th text phrase to a certain visual region.

According to the aforementioned manner, we can calculate sample-wise fine-grained alignment score matrix $\mathcal{H}\in\mathbb{R}^{B\times B}$. Subsequently, we use triplet loss with hard negative mining to maximize the scores of positive pairs and minimize the scores of negative pairs, \textit{i.e.}, we optimize 
\begin{equation}
\begin{split}
        \mathcal{L}_{fine} = \frac{1}{B}\sum_{i=1}^B\max\big(\mathcal{H}(\mathcal{I}_i, \mathcal{T}^*) -\mathcal{H}(\mathcal{I}_i, \mathcal{T}_i) + \alpha, 0\big) \\
        +  \frac{1}{B}\sum_{i=1}^B\max\big(\mathcal{H}(\mathcal{I}^*, \mathcal{T}_i) -\mathcal{H}(\mathcal{I}_i, \mathcal{T}_i) + \alpha, 0\big)
\end{split}
\end{equation}
where $\alpha$ is the margin between positive pairs and negative pairs, $\mathcal{T}^*$ and $\mathcal{I}^*$ denote the hard negative samples of $i$-th visual or textual sample, respectively.

\subsection{Loss Function}
The total loss function can be formulated as 
\begin{equation}
    \mathcal{L} = \mathcal{L}_{cls} + \lambda(\mathcal{L}_{global} + \mathcal{L}_{fine})
\end{equation}
where $\lambda$ is to balance the classification loss and alignment loss.

\section{Experiments}
\subsection{Implementation Details}
The encoders in the DAF-Net are initialized with the pre-trained BLIP backbone \cite{li2022blip}, leveraging its strong image-text alignment capability. Both the textual and visual SFAM consist of 2 stacked transformer blocks, with 32 learnable query vectors. The model is trained for 10 epochs on 4 NVIDIA RTX 5090 GPUs with a batch size of 64. Input face attack images are resized to $224\times224$, while textual descriptions are dynamically padded based on the longest sequence in each mini-batch. We adopt the AdamW \cite{loshchilov2017decoupled} optimizer with a weight decay of 0.01. The learning rate follows a warmup-cosine annealing schedule: it first linearly increases from 5e-6 to 5e-5 during warmup, then gradually decays back to 5e-6. Additionally, the temperature $\tau$ in global alignment is set to 0.07, the margin $\alpha$ in fine-grained alignment is set to 0.2, and the weighting factor $\lambda$ for alignment loss is set to 0.1.

\subsection{Evaluation Metrics}
Following standard practice in face attack detection, we employ three metrics: the Average Classification Error Rate (ACER), providing a balanced measure in classification; classification Accuracy (Acc), reflecting overall prediction correctness; and the F1-score, which balances precision and recall to account for potential class imbalance. 

\begin{table}
\centering
\caption{An overview of experimental results. ``CL Text"(class-level text) denotes that the text only contains attack types; ``CG Text" (coarse-grained text) denotes the descriptions from MS-UFAD dataset; ``FG Text" (fine-grained text) denotes our generated descriptions. $\uparrow$ denotes the higher is better and vice versa. ``F1" denotes F1-Score.}
\label{tab:overall_results}
\begin{tabular}{p{0.05cm}|l|l|c|c|c}
\hline
 &Data & Network &ACER$\downarrow$ &ACC$\uparrow$&F1$\uparrow$  \\
\hline
1&\multirow{4}{*}{Vision-only}&ResNet50 &  17.10 & 86.39& 85.11  \\
2&&ViT & 19.28  &86.42  & 82.38  \\
3&&BLIP-ViT & 16.89& 86.78 & 84.65  \\
4&&BLIP-ViT+SFAM & 15.88 & 87.52 & 85.28  \\
\hline
5&CL Text& DAF-Net &15.37 &88.02&85.00\\
\hline
6&CG Text& DAF-Net& 15.24& 88.31& 85.51 \\
 \hline
7&FG Text & DAF-Net&\textbf{12.30} & \textbf{90.34}&\textbf{87.13} \\
\hline
\end{tabular}

\end{table}

\subsection{Overall Results}
The overall experimental results are shown in TABLE \ref{tab:overall_results}. We mainly evaluate two categories of approaches: vision-only methods and text-guided methods. Vision-only methods encompass classical vision models (\textit{e.g.}, ResNet50 \cite{he2016deep}, ViT\cite{dosovitskiy2020image}) and the vision encoder of BLIP \cite{li2022blip}, denoted as BLIP-ViT. Additionally, we also train BLIP-ViT with the proposed SFAM, denoted as BLIP-ViT+SFAM, to further verify that performance gains stem from the textual descriptions rather than SFAM. Among text-guided methods, we compare multi-granularity textual inputs. The use of class-level text (\textit{e.g.}, \textit{"This is a deepfake face"}) yields only marginal improvements over vision-only methods. In contrast, coarse-grained textual descriptions from the MS-UFAD dataset are also evaluated and the results outperform both vision-only methods and the simpler class-level text approach.

However, our proposed method, which leverages fine-grained textual descriptions, achieves substantial improvements over all the aforementioned approaches. As summarized in TABLE \ref{tab:overall_results}, our method attains the lowest ACER and the highest ACC and F1-score. It outperforms the approach using coarse-grained textual descriptions from the MS-UFAD dataset by 2.94\% in ACER, 2.03\% in ACC, and 1.62\% in F1-score, respectively

This superior performance can be attributed to the fine-grained textual descriptions which describe the artifacts of face attack images in detail. Instead of solely using an image or coarse-grained descriptions, our approach aligns visual regions with fine-grained textual depictions well that enhances the model’s capacity to discriminate subtle and diverse manipulation traces, thereby learning a more generalizable and semantically grounded representation of forgery features. In summary, we can conclude that fine-grained textual guidance is highly effective for improving generalization in face attack detection.

\subsection{Ablation Study}
\subsubsection{The Impact of Alignment}
As shown in the TABLE. \ref{tab:ablation_studies}-1, removing the global alignment loss leads to a performance degradation, with ACER increasing from 12.30 to 14.76 and ACC/F1-score dropping to 88.59/85.88. A more significant decline is observed when the fine-grained alignment loss is ablated as shown in the TABLE. \ref{tab:ablation_studies}-2, resulting in a higher ACER of 15.65 and lower ACC/F1-score of 87.60/83.14, further highlighting the importance of the fine-grained textual information to the model's performance. 

\subsubsection{Semantic Forgery Aggregation Module}
As shown in the TABLE. \ref{tab:ablation_studies}-3\&4\&5, the significance of the Semantic Forgery Aggregation Module (SFAM) is demonstrated. Ablating the textual SFAM leads to a considerable performance degradation, and removing the visual SFAM shows a comparable negative impact. Interestingly, when both SFAMs are removed simultaneously, the performance is better than when only one is removed.  We speculate that removing a single SFAM implies that alignment operates at the region-word or patch-phrase level. This granularity mismatch leads to a less effective cross-modal representation compared to patch-word alignment, suggesting that maintaining comparable granularity between visual and textual tokens is crucial for fine-grained cross-modal alignment. 

However, as discussed in Section \ref{SFAM}, the granularity of patch-word alignment is excessively fine for face attack detection, where region-phrase alignment is of greater utility. As shown in TABLE. \ref{tab:ablation_studies}-5\&6, The region-phrase alignment implemented via both visual and textual SFAMs achieves superior performance compared to patch-word alignment with relative improvements of 4.61\% in ACER, 3.84\% in accuracy, and 3.17\% in F1-score.

\begin{table}
\centering
\caption{The experimental Results of Ablation Studies}
\label{tab:ablation_studies}
\begin{tabular}{l|l|l|c|c|c}
\hline
& Exp & Settings &ACER$\downarrow$ &ACC$\uparrow$&F1$\uparrow$  \\
\hline
1&\multirow{2}{*}{Alignment}& w/o $\mathcal{L}_{global}$ &   14.76 &88.59 &85.88   \\
2&& w/o $\mathcal{L}_{fine}$ & 15.65  &87.60 &83.14 \\
\hline
3&\multirow{3}{*}{SFAM} & w/o textual SFAM & 17.60 &85.85&83.99\\
4& &w/o visual SFAM & 17.63 & 86.13&83.96\\
5& &w/o both & 16.91&86.50&83.96\\
\hline
6& Ours &-&\textbf{12.30} &\textbf{90.34}&\textbf{87.13}\\
\hline
\end{tabular}

\end{table}



\begin{figure}[ht]
    \centering
    \includegraphics[width=0.85\linewidth]{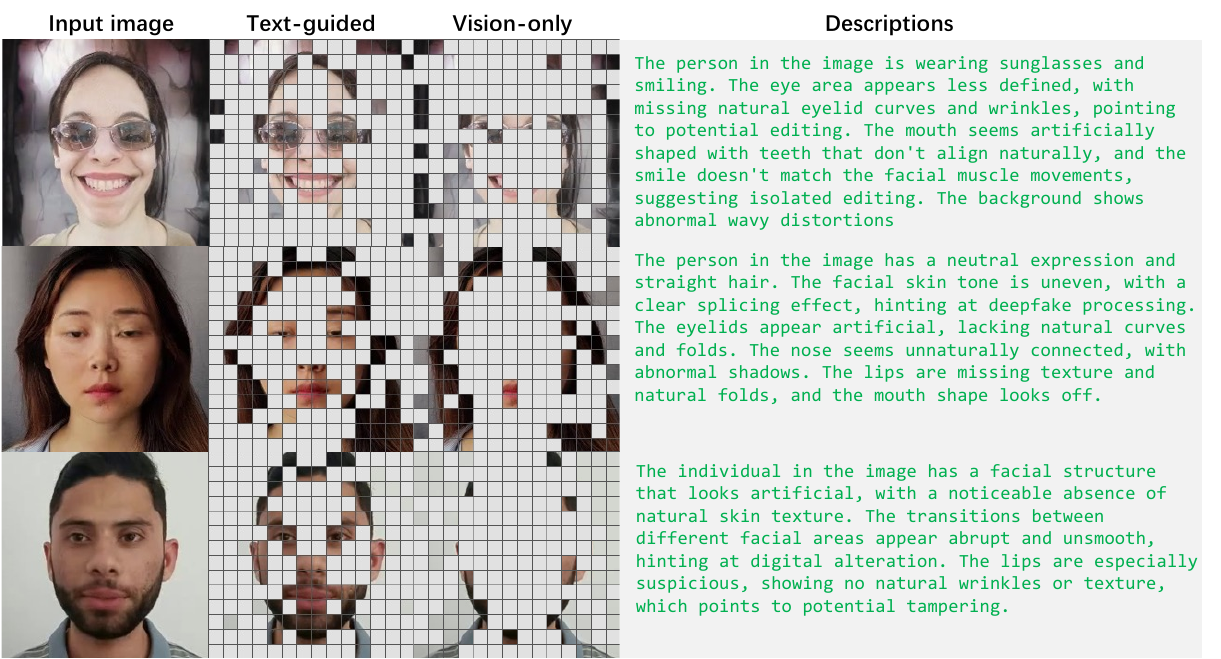}
    \caption{A visual comparison between our method (second column) and vision-only approach (third column; here, BLIP-ViT+SFAM is used). }
    \label{fig:visual}
\end{figure}
\subsection{Visualization Analysis}
We utilize the attention weights between the learnable query vectors and the input visual patches to select the top-$k$ patches ($k=64$ here) with the highest weights. As shown in Fig. \ref{fig:visual}, the vision-only method exhibits irregular attention patterns and lacks interpretability. In contrast, guided by fine-grained textual descriptions, our method more accurately focuses on the regions that are actually manipulated, which aligns with the content described in the corresponding descriptions.

\section{Conclusion}
In conclusion, we propose to enhance the face attack detection model by incorporating fine-grained textual descriptions. We enrich the MS-UFAD dataset with 8 million fine-grained annotations and propose DAF-Net, which learns more semantically meaningful forgery representations through cross modal alignment. Comprehensive experiments demonstrate that our method significantly improves performance, offering an effective solution for building more secure and robust face attack detection systems.

\bibliographystyle{IEEEtran}
\bibliography{IEEEexample}

\end{document}